\documentclass{article}

\usepackage[preprint,nonatbib]{neurips_2021}

\usepackage[utf8]{inputenc}
\usepackage{hyperref}

\usepackage[inline]{enumitem}
\usepackage{bbm}
\usepackage{graphicx} 
\usepackage{booktabs}
\usepackage{amsmath,amssymb,amsthm,dsfont}
\usepackage{mathtools}
\usepackage{caption,subcaption}
\usepackage{tikz,pgfplots,filecontents}
\usepackage{multirow}
\usepackage{appendix}
\usepackage{algorithm}
\usepackage{algorithmic}
\usepackage[numbers]{natbib}

\mathtoolsset{showonlyrefs=true,showmanualtags=true}
\usepackage{bm}

\newcommand{\inx}{\ensuremath{\mathcal{X}}}
 
\newcommand{\inz}{\ensuremath{\mathcal{Z}}}

\newcommand{\pp}[1]{\ensuremath{\mathbb{#1}}}     

\newcommand{\hbsph}{\ensuremath{\mathcal{H}}}   

\newcommand{\rd}[1]{\mathbb{R}^{#1}}              
\newcommand{\rr}{\mathbb{R}} 		         
\newcommand{\ep}{\mathbb{E}}                     

\newcommand{\dd}{\, \mathrm{d}}

\newcommand{\x}{\ensuremath{\mathbf{x}}}

\newcommand{\pto}{\overset{\mathrm{p}}{\to}}
\newcommand{\dto}{\overset{\mathrm{d}}{\to}}

\newcommand{\argmin}{\operatornamewithlimits{argmin}}
 
\usetikzlibrary{shapes,decorations,arrows,calc,arrows.meta,fit,positioning,decorations.pathreplacing}
\newtheorem{theorem}{Theorem}

\newtheorem{lemma}{Lemma}
\newtheorem{remark}{Remark}

\newtheorem{assumption}{Assumption}
\newtheorem{definition}{Definition}

\usepackage{minitoc}

\usepackage{color}

\title{Instrument Space Selection for Kernel Maximum Moment Restriction}

\newcommand{\anusym}{ \S}
\newcommand{\utsym}{ \dag}
\newcommand{\mpisym}{ \ddag}

\author{%
  Rui Zhang\textsuperscript{\anusym} \hspace{0.7cm}
  Krikamol Muandet\textsuperscript{\mpisym} \hspace{0.7cm}
  Bernhard Sch\"olkopf\textsuperscript{\mpisym} \hspace{0.7cm}
  Masaaki Imaizumi\textsuperscript{\utsym}
  \\
  \begin{tabular}{ccc}
  \textsuperscript{\anusym} Australian National University
  & \textsuperscript{\mpisym}MPI for Intelligent Systems & \textsuperscript{\utsym}University of Tokyo Tokyo
  \\
  Canberra, Australia & T\"ubingen, Germany& Tokyo, Japan
  \end{tabular}
}

\begin{document}

\maketitle

\begin{abstract}
    
    Kernel maximum moment restriction (KMMR) recently emerges as a popular framework for instrumental variable (IV) based conditional moment restriction (CMR) models with important applications in conditional moment (CM) testing and parameter estimation for IV regression and proximal causal learning.
    The effectiveness of this framework, however, depends critically on the choice of a reproducing kernel Hilbert space (RKHS) chosen as a space of instruments. 
    In this work, we presents a systematic way to select the instrument space for parameter estimation based on a principle of the least identifiable instrument space (LIIS) that identifies model parameters with the least space complexity. 
    Our selection criterion combines two distinct objectives to determine such an optimal space: (i) a test criterion to check identifiability; (ii) an information criterion based on the effective dimension of RKHSs as a complexity measure. 
    We analyze the consistency of our method in determining the LIIS, and demonstrate its effectiveness for parameter estimation via simulations.
\end{abstract}

\section{Introduction}

The instrumental variable (IV) based conditional moment restriction (CMR) models \citep{Newey93:CMR,ai03Efficient,Dikkala20:Minimax} have a wide range of applications in causal inference, economics, and finance modeling, where for correctly-specified models the conditional mean of certain functions of data equals zero almost surely.
This kind of models also appear in Mendelian randomization, a technique in genetic epidemiology that uses genetic variation to improve causal inference of a modifiable exposure on disease \citep{davey2003mendelian,Burgess17review}. 
Rational expectation models \citep{muth1961rational}, widely-used in macroeconomics, measures how available information is exploited to
form future expectations by decision-makers as conditional moments \citep{muth1961rational}. 
Furthermore, CMRs have also gained popularity in the community of causal machine learning, leading to novel algorithms such as generalized random forests \citep{athey2019generalized}, double/debiased machine learning \citep{chernozhukov2018double} and nonparametric IV regression \citep{Bennett19:DeepGMM,Muandet20:DualIV,zhang2020maximum}; see also related works therein, as well as in offline reinforcement learning \citep{liao2021instrumental}.

Learning with CMRs is challenging because it implies an infinite number of unconditional moment restrictions (UMRs).
Although the asymptotic efficiency of the instrumental variable (IV) estimator can in principle improve when we add more moment restrictions, it was observed that the excessive number of moments can be harmful in practice \cite{Andersen96:GMM} because its finite-sample bias increases with the number of moment conditions \cite{Newey04Higher}.
Hence, traditional works in econometrics often select a finite number of UMRs for estimation based on the generalized method of moments (GMM) \citep{hansen1982large,hall2005generalized}. 
Unfortunately, an adhoc choice of moments can potentially lead to a loss of efficiency or even a loss of identification \citep{Dominguez04:GMM}.
For this reason, subsequent works advocate an incorporation of all moment restrictions simultaneously in different ways such as the method of sieves \citep{DeJong1996Bierens,Donald2003Empirical} and a continuum of moment restrictions \citep{Carrasco2000:Continuum,Carrasco07:LIP, carrasco2012regularization,Carrasco14:Efficiency}, among others.
Despite the progress, the question of moment selection in general remains open. 

A recent interest to model the CMR using a reproducing kernel Hilbert space (RKHS) \citep{Muandet20Kernel,Dikkala20:Minimax} and deep neural networks \citep{Lewis18:AGMM,Bennett19:DeepGMM} opens up a new possibility to resolve the selection problem with modern tools in machine learning. 
In this work, we focus on the RKHS approach where the CMR is reformulated as a minimax optimization problem whose inner maximization is taken over functions in the RKHS. 
This framework is known as a \textit{kernel maximum moment restriction} (KMMR).
An advantage of the KMMR is that one can get a closed-form solution to the maximization problem, which is related to a continuum generalization of GMM \citep{Carrasco2000:Continuum,carrasco2012regularization}.
Furthermore, it has been shown that an RKHS with a specific type of kernels is sufficient to model the CMR; see \citet[Theorem 3.2]{Muandet20Kernel} and \citet[Theorem 1]{zhang2020maximum}.
Hence, in this context, the moment selection problem becomes the \emph{kernel selection} problem, which is also a long-standing problem in machine learning.
Besides, the KMMR can be viewed as an approximate dual of the known two-stage regression procedure \citep{Muandet20:DualIV, Liao20:NeuralSEM}.
KMMRs have been applied successfully to IV regression \citep{zhang2020maximum}, proximal causal learning \citep{Mastouri2021Proximal} and condition moment testing \citep{Muandet20Kernel}. 
Nevertheless, all of them employed a simple heuristic to select the kernel function, e.g., median heuristic, which limits the full potential of this framework. 

\textbf{Our contributions.}
In this paper, we aim to address the kernel selection problem for the KMMR framework. We focus on the IV estimator and term our problem the \textit{kernel} or \textit{RKHS instrument space selection}, because the RKHS functions as a space of instruments. We define an optimal instrument space, named \textit{least identifiable instrument space} (LIIS), which has the identifiability for the true model parameters and the least  complexity. To determine LIIS in practice, we propose an approach based on a combination of two criteria: (i) the identification test criterion (ITC) to test the identifiability of instrument spaces and (ii) the kernel effective information criterion (KEIC) to select the space based on its complexity. 
Our method has the following advantages: \textbf{(a)} compared with the higher-order asymptotics based methods \citep{donald2001choosing,donald2009choosing}, our approach is easy to use and analyze, and can filter invalid instrument spaces; \textbf{(b)} our method is a combination of several information criteria, therefore compensating shortcomings of individual criteria.
Moreover, we analyze the consistency of our method on selection of the LIIS and we show in the simulation experiments that our method effectively identifies the LIIS and improves the performance on parameter estimation for IV estimators. To the best of our knowledge, we do not find any other method achieve all of these on the kernel instrument space selection.

\section{Preliminaries}
\label{sec:mmr}


\subsection{Conditional moment restriction (CMR)}

Let $(X,Z)$ be a random variable taking values in $\inx\times\inz$ and $\Theta$ a parameter space.
A conditional moment restriction (CMR) \citep{Newey93:CMR,ai03Efficient} can then be expressed as 
\begin{equation}\label{eq:cmr}
    \text{CMR}(\theta_0) = \pp{E}[\varphi_{\theta_0}(X)\,|\,Z] = \mathbf{0}, \quad P_Z-\text{almost surely (a.s.)}
\end{equation}
for the true parameter $\theta_0\in\Theta$. 
The function $\varphi_{\theta}(X)$ is a problem-dependent generalized residual function in $\rd{q}$ parameterized by $\theta$. 
Intuitively, the CMR asserts that, for correctly specified models, the conditional mean of the generalized residual function is almost surely equal to zero. 
Many statistical models can be written as \eqref{eq:cmr} including nonparametric regression models where $X=(\tilde{X},Y), Z=\tilde{X}$ and $\varphi_{\theta}(X) = Y - f(\tilde{X};\theta)$; conditional quantile models where $X=(\tilde{X},Y), Z=\tilde{X}$, and $\varphi_{\theta}(X) = \mathbbm{1}\{Y < f(\tilde{X};\theta)\} - \tau$ for the target quantile $\tau\in[0,1]$; and  IV regression models where $X=(\tilde{X},Y)$, $Z$ is an IV, and $\varphi_{\theta}(X) = Y - f(\tilde{X};\theta)$.

\vspace{-5pt}
\paragraph{Maximum moment restriction (MMR).}
An important observation about the CMR \eqref{eq:cmr} is that it implies a continuum of unconditional moment restriction (UMR) \citep{Carrasco2000:Continuum,Lewis18:AGMM,Bennett19:DeepGMM}:
$\pp{E}[\varphi_{\theta_0}(X)^\top h(Z)] = 0$ for all measurable functions $h\in \hbsph$ where $\hbsph$ is a space of measurable functions $h:\inz \to \rr^q$. 
We refer to $\hbsph$ as an \textit{instrument space}.
Traditionally, inference and estimation of $\theta_0$ can be performed, for example, via a generalized method of moment (GMM) of UMR based on a specific subset of $\hbsph$ \citep{hall2005generalized}. 
Consequently, the choice of subset of $\hbsph$ plays an important role in parameter estimation for the conditional moment models.
In this paper, we discuss on the optimal instrument space $\hbsph$ which is developed based on an equivalent definition of UMR, called the maximum moment restriction (MMR) \citep{Lewis18:AGMM,Muandet20Kernel,zhang2020maximum}, as follows:
\begin{equation}\label{eq:umr_h}
    R_{\hbsph}(\theta_0)\coloneqq \sup_{h \in \hbsph }\pp{E}^2[\varphi_{\theta_0}(X)^\top h(Z)] = 0.
\end{equation}
Note that the MMR $R_{\hbsph}$ depends critically on the choice of an instrument space $\hbsph$. 
In this paper, we focus exclusively on the IV regression models such that $\varphi_{\theta}(X) = Y - f(\tilde{X};\theta) \in \rr$ and $\hbsph$ is a real-valued function space. 
We defer applications of our method in other scenarios to future work.

\subsection{Kernel maximum moment restriction (KMMR)}


In this work, we focus on $R_{\hbsph}(\theta)$ when an instrument space $\hbsph$ is a reproducing kernel Hilbert space (RKHS) associated with the kernel $k:\inz\times\inz\to\mathbb{R}$ \cite{Aronszajn50:Reproducing, Scholkopf01:LKS, Berlinet04:RKHS}.

\vspace{-5pt}
\paragraph{Reproducing kernel Hilbert space (RKHS).} 
Let $\hbsph$ be a reproducing kernel Hilbert space (RKHS) of functions from $\inz$ to $\mathbb{R}$ with $\langle \cdot,\cdot\rangle_{\hbsph}$ and $\|\cdot\|_{\hbsph}$ being its inner product and norm, respectively. 
Since for any $z\in\inz$ the linear functional $h \mapsto h(z)$ is continuous for $h\in\hbsph$, it follows from Riesz representation theorem \citep{riesz1909:representation} that there exists, for every $z\in\inz$, a function $k(z,\cdot) \in \hbsph$ such that $h(z) = \langle h, k(z,\cdot)\rangle_{\hbsph}$ for all $h\in\hbsph$. 
This is generally known as a reproducing property of $\hbsph$ \citep{Aronszajn50:Reproducing,Scholkopf01:LKS}.
We call $k(z,z') := \langle k(z,\cdot), k(z',\cdot)\rangle_{\hbsph}$ a reproducing kernel of $\hbsph$.
The reproducing kernel $k$ is unique (up to an isometry) and fully characterizes the RKHS $\hbsph$ \cite{Aronszajn50:Reproducing}. 
Examples of commonly used kernels on $\rr^d$ include Gaussian RBF kernel $k(z,z') = \exp(-\|z-z'\|^2_2/2\sigma^2)$ and Laplacian kernel $k(z,z') = \exp(-\|z-z'\|_1/\sigma)$ where $\sigma > 0$ is a bandwidth parameter.
For the detailed exposition on kernel methods and RKHS, see, e.g., \citep{Scholkopf01:LKS,Berlinet04:RKHS,Micchelli05:RKHS,Muandet17:KME}.

By representing $\hbsph$ in \eqref{eq:umr_h} with the RKHS, \citet[Theorem 3.3]{Muandet20Kernel} showed that $R_{\hbsph}(\theta)$ has a closed-form expression after introducing an Ivanov regularization \citep{Ivanov02:Ill-posed} $\|h\|=1$ to remove the scale effect of instruments:
\begin{align}\label{eq:mmr_k}
    R_{\hbsph}(\theta) := \sup_{h \in \hbsph, \| h\|=1}\pp{E}^2[\varphi_{\theta}(X) h(Z)] = \pp{E}[\varphi_{\theta}(X) k(Z,Z')\varphi_{\theta}(X')],
\end{align}
where $(X',Z')$ is an independent copy of $(X,Z)$. 
Given i.i.d data $\{(x_i,z_i)\}_{i=1}^n$, we define its empirical analogue $ \hat{R}_{\hbsph}({\theta}) \coloneqq \ep_n[\varphi_{\theta}(X)K(Z,Z)\varphi_{\theta}(X)]$ and its minimizer $\hat{\theta} = \argmin_\theta \hat{R}_{\hbsph}({\theta})$.

We focus on this expression in spite of a similar quadratic expression following from a Tikhonov regularization on $h$ \citep[Eqn.~(10)]{Dikkala20:Minimax}. 
It is instructive to observe that the MMR \eqref{eq:mmr_k} resembles the optimally-weighted GMM formulation of \citet{Carrasco2000:Continuum}, but without the re-weighting matrix; see, also, \citet[Sec. 6]{Carrasco07:LIP} and \citet[Sec. 3]{zhang2020maximum}. 
While the optimally-weighted GMM (OWGMM) was originally motivated by the asymptotic efficiency theory in a parametric setting \citep{Carrasco14:Efficiency}, the need to compute the inverse of the parameter-dependent covariance operator can lead to more cumbersome estimation \citep{carrasco2012regularization} and poor finite-sample performance \citep{bond2002finite}.
Hence, we will consider $R_{\hbsph}(\theta)$ throughout for its simplicity.

The following result adapted from \citet[Theorem 1]{zhang2020maximum} and \citet[Theorem 3.2]{Muandet20Kernel} guarantees that KMMR has the same roots as those of CMR \eqref{eq:cmr}. 

\begin{theorem}[Sufficiency of the instrument space]\label{thm:sufficiency}
Suppose that $k$ is continuous, bounded (i.e., $\sup_{z \in \inz}\sqrt{k(z,z)} < \infty$) and satisfies the condition of integrally strictly positive definite (ISPD) kernels, i.e., for any function $g$ that satisfies $0 < \|g\|_2^2 < \infty$, we have $\iint_{\inz} g(z)k(z,z')g(z') \dd z \dd z'>0$. 
Then, $R_{\hbsph}(\theta) = 0$ if and only if $\text{CMR}(\theta) = \mathbb{E}[\varphi_{\theta}(X)\,|\,Z] = \mathbf{0}$ for $P_Z$-almost all $z$.
\end{theorem}

In other words, Theorem \ref{thm:sufficiency} implies that it is sufficient to restrict $\hbsph$ in \eqref{eq:umr_h} to the RKHS associated with the ISPD kernel. However, it does not guarantee the optimality of $\hbsph$ as an instrument space.

\section{Least identifiable instrument space (LIIS)}
\label{sec:least-identify}

The choice of an instrument space is critical for KMMR.
If an instrument space is excessively small, KMMR loses an identification power, i.e., another parameter $\theta \neq \theta_0$ can satisfy the KMMR condition \eqref{eq:mmr_k}, hence it is impossible in principle to achieve a consistent estimator to $\theta_0$.
The scenario is often referred to as the \textit{under-identification problem} \citep[Chapter 2.1]{hall2005generalized}.
In contrast, an excessively large instrument space increases an error of estimators with finite samples.
It is proved that a mean squared error (MSE) of an estimator with MMR is increased by the size of an instrument space \citep{breusch1999redundancy,hall2007information,cheng2015select,Dikkala20:Minimax}.
Specifically, for example, Theorem 1 in \cite{Dikkala20:Minimax} showed that the MSE of their estimator has an upper bound which increases in a critical radius of an instrument space (defined by an upper bound of its Rademacher complexity).
Therefore, it is important to avoid making instrument spaces excessively large in order to reduce errors.

Unfortunately, the instrument space selection problem cannot be solved by a straightforward cross-validation (CV) procedure, because the loss function \eqref{eq:mmr_k} depends on an instrument space itself.
For example, we can rewrite KMMR as
$R_{\mathcal{H}}(\theta) = \sum_{i\geq 1}\lambda_i \mathbb{E}[\varphi_{\theta}(X)\phi_i(Z)]^2$ by the Mercer's decomposition of a corresponding kernel $k(z,z') = \sum_{i\geq 1}\lambda_i\phi_i(z)\phi_i(z')$ with eigenfunctions and eigenvalues  $\{\phi_i(\cdot), \lambda_i\}_{i\geq 1}$.
Due to the dependence, CV always select an excessively small instrument space which makes $R_{\mathcal{H}}$ always small.

\tikzset{
    -Latex,auto,node distance =1 cm and 1 cm,semithick,
    state/.style ={ellipse, draw, minimum width = 0.7 cm},
    point/.style = {circle, draw, inner sep=0.04cm,node contents={}},
    bidirected/.style={Latex-Latex,dashed},
    el/.style = {inner sep=2pt, align=left, sloped}
}
\begin{figure}[t!]
    \centering
    \begin{tikzpicture}
        \node (p2) at (-3,0) [label=65:$\hbsph^*$,point]; %
        \node (p0) at (0,0) [point];
        \node (a) at (3,0) [right] {$\hbsph_j$} ;
        \node[below] at (-0.3,-1.15) {weak-};

        \path (p0) edge [-Latex,line width=1pt] node [below] {over-} (3,0);
        \path (p2) edge [-,line width=1pt] node [below] {identification} (p0);
        \path (-6,0) edge [-,line width=1pt]  node [below] {under-}    (p2);
        
    \draw[->] (-5.8, -1.6) -- (-3.2, -1.6) node[right] {};
    \draw[->] (-4, -1.8) -- (-4, -0.5) node[below left] {$R$};
    \draw[scale=1, domain=-5.5:-3.5, smooth, variable=\x, blue,-] plot ({\x}, {1.5*(\x+4.8)*(\x+4.8)*(\x<-4.5)+1.5*(\x+4.2)*(\x+4.2)*(\x>=-4.5)-1.6});
    
    \draw[->] (-2.8, -1.6) -- (-0.2, -1.6) node[right] {};
    \draw[->] (-1, -1.8) -- (-1, -0.5) node[below left] {$R$};
    \draw[scale=1, domain=-2.5:-0.5, smooth, variable=\x, black, dashed,-] plot ({\x}, {0.7*(\x+1.5)*(\x+1.5)-1.6});
    \draw[scale=1, domain=-2.5:-0.5, smooth, variable=\x, blue,-] plot ({\x}, {1.2*(\x+1.5)*(\x+1.5)-1.6});
    \draw[->] (0.5, -1.6) -- (2.8, -1.6) node[right] {$\theta$};
    \draw[->] (2, -1.8) -- (2, -0.5) node[below left] {$R$};
    \draw[scale=1, domain=0.5:2.5, smooth, variable=\x, blue,-] plot ({\x}, {(\x-1.5)*(\x-1.5)-1.5});
    \end{tikzpicture}
    \caption{Illustration of instrument spaces and identifiability. Instrument spaces are located on the top horizontal axis with higher complexity on the right. $H^{*}$ denotes the LIIS. Spaces are categorized into three classes: under-, identification and over-. Each category has an example below. Under-identification has the MMR value $R=0$ at multiple model parameters $\theta$'s; identification: unique; over-identification: none; weak identification: unique but $R\approx 0$ at multiple $\theta$'s.}
    \label{fig:space_h_ident}
\end{figure}
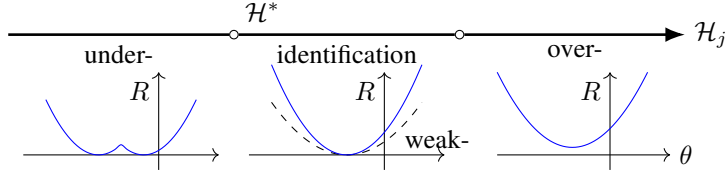

\textbf{The proposed optimality.}
To resolve this issue, we may consider candidates of instrument spaces $\hbsph_1,...,\hbsph_M$.
For example, we consider RKHSs by the Gaussian kernel with different lengthscale parameters.
We introduce assumptions on the identification of $\theta_0$ and identifiability of $\hbsph_i$.
\begin{assumption}[Global identification]\label{asmp:global_ident}
    There exists an unique $\theta_0 \in \Theta$ satisfying $\mathrm{CMR}(\theta_0)=0$.
\end{assumption}
\begin{assumption}[Identifiability] \label{asmp:identifiable}
    There exists  non-empty index set $\mathcal{M} \subset \{1,2,...,M\}$ such that for any $m \in \mathcal{M}$, there is $\theta_0$ \textit{uniquely} satisfies $R_{\hbsph_m}(\theta_0)=0$.
\end{assumption}
These assumptions guarantee that there is at least one instrument space from the candidates identifying the unique solution $\theta_0$ of the CMR problem \eqref{eq:cmr}.
We, then, define a notion of optimality of instrument spaces. 
Let $\Omega(\hbsph)$ be a complexity measure of $\hbsph$, which is specified later.
\begin{definition}[Least Identifiable Instrument Space (LIIS)]\label{def:least_identifiable}
    A \textrm{least identifiable instrument space} $\hbsph^*$ is an identifiable instrument space with the least complexity, i.e. $\hbsph^* := \argmin_{\hbsph_{j}: j \in \mathcal{M}} \Omega(\hbsph_j)$.
\end{definition}
The notion of LIIS is designed to satisfy the two requirements: the identifiability to $\theta_0$, and the less complexity to reduce an estimation error.
This optimality is different from the way with test errors, such as CV.
We provide an illustration of the concepts in Figure~\ref{fig:space_h_ident}.



\section{Least identification selection criterion (LISC)}
\label{sec:selection-criteria}


We propose a method to find LIIS from finite samples, named the \textit{least identification selection criterion} (LISC), by developing several criteria necessary for LIIS and combining them.
The developed criteria are as follow: an identification test criterion (ITC) and a kernel effective information criterion (KEIC). 
We first introduce the overall methodology and then explain these criteria. 

Based on the ITC and KEIC, we propose a simple two-step procedure to select the optimal space. We first select a set of identifiable spaces via the ITC and then choose an identifiable space with the least KEIC. In case there is no identifiable instrument space determined when e.g. neural networks are employed, we minimize the ratio of the ITC and KEIC to select LIIS:
\begin{align*}
    \textstyle \hat{\hbsph} = \argmin_{\hbsph_j: j = 1,...,M} \mathrm{KEIC}(\hbsph_j)/\mathrm{ITC}(\hbsph_j).
\end{align*}
Note that the ratio based method follows the spirit of LIIS and of our two-step procedure.

\subsection{Criterion 1: identification test criterion (ITC)}

We first develop a method to study identifiability of instrument spaces, by validating whether a minimizer $\theta$ of $R_\hbsph(\theta)$ is unique. 
The uniqueness is verified by examining a rank of a Hessian-like matrix from $R_\hbsph(\theta)$, which is $F_\hbsph(\theta_*):=\ep[u_{\theta_*}(S)]$ where $u_{\theta}(S) \coloneqq \nabla_{\theta} \phi_{\theta^*}(X)k(Z,Z')\nabla_{\theta} \phi_{\theta^*}(X')$ with $S\coloneqq (X,Z,X',Z')$ and $\theta_*$ is a minimizer of $R_{\hbsph}$ assumed to satisfy $R_{\hbsph}(\theta_*)=0$.
A full-rankness of the matrix $F_\hbsph(\theta_*)$ is a sufficient condition for global identification on linear models $\varphi_{\theta}$,
and local identification on nonlinear models \citep[Assumptions 2.3, 3.6]{hall2005generalized}.

\textbf{Test of full-rankness.}
We develop a statistical test for the full-rankness of $F_{\hbsph}(\theta_*)$, based on the test of ranks \citep{robin2000tests}.
With a $c$-dimensional parameter $\theta$ and a $c \times c$ matrix $F_\hbsph(\theta_*)$, we consider a null ($H_0:\mathrm{rank}(F_\hbsph(\theta)) = c-1$) and alternative ($H_1: \mathrm{rank}(F_\hbsph(\theta))=c$) hypotheses.
Let $\lambda_c$ be the smallest eigenvalue of $F_\hbsph(\theta_*)$, which is non-negative owing to the quadratic form of $F_{\hbsph}$. 
For the test, we consider an empirical analogue of $F_\hbsph(\theta_*)$ as $\hat{F}_\hbsph(\hat{\theta})= \ep_n[u_{\hat{\theta} }(S)]\coloneqq n^{-2}\sum_{i,j=1}^{n} u_{\hat{\theta} }(s_{ij})$ with $s_{ij}:=(x_i,z_i,x_j,z_j)$ and $\hat{\theta} \coloneqq\argmin_{\theta} \hat{F}_{\hbsph}(\theta)$.
Then, we consider the smallest eigenvalue $\hat{\lambda}_c\geq 0$ of $\hat{F}_\hbsph(\hat{\theta})$ and employ it as a test statistics $\hat{T} := \hat{\lambda}_c^2$.
Let $N$ be a standard normal random variable and $\Lambda$ be a fixed scalar given in Appendix \ref{app:itc}.
For a significance level $\alpha \in (0,1)$, such as $\alpha=0.05$, let $Q_{1-\alpha}$ be a $(1-\alpha)$-quantile of $N^2$.
Our test has the limiting power of one:
\begin{theorem}\label{thm:test_power}
    Assume the conditions of Theorem \ref{thm:rank-asymptotic}. The test that rejects the
    null $\mathrm{rank}(F_{\hbsph}(\theta_*))=c-1$ when $n\hat{T} > \Lambda Q_{1-\alpha}$ is consistent against
    any fixed alternative $\mathrm{rank}(F_{\hbsph}(\theta_*))=c$.
\end{theorem}

\textbf{Test criterion.}
We propose a criterion based on the test statistic $\hat{T}$.
We randomly split the dataset into two parts and use one part to compute $\hat{T}$ and the other part for computing $\hat{\Lambda}=(\hat{C} \otimes \hat{C})^\top \hat{\Omega} (\hat{C}\otimes \hat{C})$ where $\hat{C}$ is an eigenvector of $\hat{F}_\hbsph(\hat{\theta})$ corresponding to the smallest eigenvalue $\hat{\lambda}_c$, and $\hat{\Omega} = \ep_{n} [ \mathrm{vec}(u_{\hat{\theta}}(S))\mathrm{vec}( u_{\hat{\theta}}(S))^{\top}]-\ep_{n} [ \mathrm{vec}( u_{\hat{\theta}}(S))]\ep_{n}[ \mathrm{vec}( u_{\hat{\theta}}(S))]^{\top}$.
Here, $\mathrm{vec}(\cdot)$ is the vectorization operator.
Then, we define an \textit{identification test criterion} (ITC) as
\begin{align*}
    \mathrm{ITC}(\hbsph) := n \hat{T}/\hat{\Lambda}.
\end{align*}
We select an instrument space under which we can reject the null-hypothesis, namely, we select $\hbsph$ when $\mathrm{ITC}(\hbsph) > Q_{1-\alpha}$ holds.
The validity of the selection is shown in the following result:
\begin{theorem}[Consistency of ITC] \label{thm:consistency_ITC}
    Suppose that conditions of Theorem \ref{thm:test_power} and Assumption \ref{asmp:global_ident} hold,  $\Theta$ is compact, $R_{\hbsph}(\theta)$ is consistent to $\mathrm{CMR}(\theta)$, $\hat{R}_{\hbsph}(\theta)$ converges to $R_{\hbsph}(\theta)$ uniformly in probability and $R_{\hbsph}(\theta)$ is finite. Then the instrumental space selected by ITC is identifiable in probability approaching to $1$ as $n \to \infty$.
\end{theorem}

\subsection{Criterion 2: kernel effective information criterion (KEIC)}

We develop another criterion for LIIS in the spirits of the \textit{information criterion} like \citet{akaike1974new} and \citet{Andrews99:SelectGMM}.
The strategy is to estimate both elements required for LIIS: to measure the complexity of $\hbsph$ by the notion of \textit{effective dimension}, and the identifiability by the empirical loss.


\textbf{Effective dimension.} The effective dimension has been a common measure of a complexity of RKHSs \citep{Zhang2003Effective,Mendelson03performance,caponnetto2007optimal} and used to analyze the performances
of kernel methods.
To develop the notion, we consider the Mercer expansion \citep{Mercer1909functions} of $k(z,z') = \sum_{i\geq 1}\lambda_i^k\phi_i(z)\phi_i(z')$ for $q=1$ as provided in Section \ref{sec:least-identify}, with a superscript $k$ to differentiate eigenvalues of $F_{\hbsph}$ and those of the kernel $k$.
Based on \cite{caponnetto2007optimal}, a definition of the effective dimension is written as
\begin{align}\label{def:effective_dim}
    \textstyle \mathrm{E}_{k} = \left(\sum_{i=1}^{\infty} \lambda_i^k\right)\left[\sum_{i=1}^{\infty} (\lambda_i^{k})^2\right ]^{-\frac{1}{2}}. 
\end{align}
We develop an empirical estimator for $\mathrm{E}_{k}$ as
    $\mathrm{Tr}(K_{\bm z}){\mathrm{Tr}(K_{\bm z}^2)^{-\frac{1}{2}}}$,
where $[K_{\bm z}]_{ij} = k(z_i,z_j)$ is a kernel matrix and $\mathrm{Tr}(\cdot)$ is the trace operator. 
We show that its consistency as follows.
\begin{theorem} \label{thm:consistency_E}
As $n \to \infty$, $\mathrm{Tr}(K_{\bm z}){\mathrm{Tr}(K_{\bm z}^2)^{-\frac{1}{2}}} \to \mathrm{E}_k$ holds.
\end{theorem}
The effective dimension measures the complexity of the instrument space $\hbsph_k$ and meanwhile quantifies some capacity properties of the marginal measure $P_Z$. An interpretation to our definition is that the numerator counts the effective UMRs which are assigned to relatively larger eigenvalues, and the denominator regularizes the count. Effective dimensions may differ in different tasks. For example, $\sum_{i} \lambda_i(\lambda_i+\alpha)^{-1}$ is considered in least square regression \citep{Zhang2003Effective,caponnetto2007optimal}, where $\alpha>0$ is a regularizer parameter, and \citet{lopes2011more} interprets $(\sum_{i} \lambda_i)^2/(\sum_{i=1}^{\infty} \lambda_i^2)$ as the effective dimension of the covariance matrix.

\textbf{Information criterion.} 
We then develop an information criterion with the notion of the effective dimension.
The key idea is to add the estimated effective dimension as a penalty term for an empirical KMMR risk. Note that the ITC is developed based on the existence assumption of  parameters satisfying the KMMR, while the assumption may not hold always. Therefore, it is necessary to test the existence by the empirical risk. We propose a \textit{kernel effective information criterion} (KEIC) on  $\hbsph$ as below, which is an analogy of a standard notion of information criteria such as BIC (Bayesian Information Criterion) \citep{schwarz1978estimating}.
\begin{align*}
    \mathrm{KEIC}(\hbsph) := n\hat{R}_\hbsph(\hat{\theta}) + \mathrm{Tr}(K_{\bm z}){\mathrm{Tr}(K_{\bm z}^2)^{-1/2}} \log n.
\end{align*}
\begin{remark}
    Given a set of valid and invalid instrument spaces, the KEIC filters invalid ones in probability approaching 1 as $n \to \infty$, since they don't have zero risks, i.e., $R_{\hbsph} > 0$, and the first term in the KEIC increases faster than the second term.
\end{remark}

\begin{theorem}[Consistency of KEIC] \label{thm:keic}
    Suppose a function space $\hbsph^{l}$ satisfying Assumption \ref{asmp:identifiable} uniquely exists  and Assumption \ref{asmp:global_ident} holds.
Given a set of identifiable spaces $\hbsph_i$, $i\in \mathcal{M}$, then $\mathrm{P}( \tilde{\hbsph} = \hbsph^l ) \to 1$ as $n \to \infty$ where  $\tilde{\hbsph} = \argmin_{ \hbsph_{j}: j \in \mathcal{M}} \mathrm{KEIC}(\hbsph_j)
$.
\end{theorem}
\begin{remark}[Consistency of two-step procedure]
Suppose that conditions of Theorem \ref{thm:keic} hold.
Then consistency of the two-step selection procedure for the LIIS holds if the selected space by ITC is identifiable in probability approacing to 1 as $n \to \infty$.
\end{remark}

\section{Related work}

The problems of moment and instrument selection have a long history in econometrics \citep{Newey93:CMR,hall2005generalized,Hall15Econometricians}.
While both problems often involve the GMM estimator, 
the latter focuses on the IVs and the corresponding estimator is referred to as the \emph{generalized instrumental variable} (GIV) estimator \citep{Hansen1982GIV}. 
In general, existing selection criteria can be summarized into three broad categories: 
(i) large sample or first-order asymptotic property based criteria, 
(ii) finite sample or higher-order asymptotic property based criteria, and 
(iii) information criteria. 

The first category of selection methods, which was popular in the 1970s-1980s, treats an asymptotic efficiency of the resulting estimator as the most desirable criterion; see \citep{Newey93:CMR} for a review. 
However, this kind of criteria may not guarantee good finite sample properties as they incur large biases in practice \citep{morimune1983approximate,donald2001choosing}. Thus, subsequent work gradually turned to the second category which aims to improve the finite sample precision of parameter estimation. 
\citet{donald2001choosing} proposed an instrument selection criterion based on a second-order asymptotic property, i.e., a Nagar type approximation \citep{nagar1959bias} to the bias of linear causal models. 
\citet{Newey04Higher} explored higher-order asymptotic properties of GMM and generalized empirical likelihood estimation, which was later applied by \citet{donald2009choosing} to developing the instrument selection for non-linear causal models. Interestingly shown by \citet{Newey04Higher}, many bias-correction methods implicitly improves higher-order asymptotic efficiency. 
Moreover, the idea of improving finite sample biases is highly related to the cross validation (see e.g. \citet[pp. 1165]{donald2001choosing} and \citet[section 4]{carrasco2012regularization}), which has been used for different targets like selection of the weight matrix of GMM \citep{carrasco2012regularization} and of regularization parameters in the GMM estimation \citep{zhang2020maximum}. 
Nonetheless, higher-order asymptotic properties often rely on complicated theoretical analyses and its practical performance is sensitive to the accuracy of the empirical approximation and noise in data \citep{Ghosh94:Higher}; the empirical approximation is also often computationally heavy (see, e.g., \citet{donald2009choosing}). Additionally, this category of methods requires a strong assumption that all instrument candidates are valid \citep{donald2001choosing,donald2009choosing}.
Therefore, it is desirable to seek a theoretically simpler selection method which remains robust and easy to use in practice. 
The last category, to which our method also belongs, relies on the information criterion. 
\citet{Andrews99:SelectGMM} previously proposed an orthogonality condition based criterion. Specifically, the method selects a maximal number of valid moments by minimizing the objective of the GMM estimation and maximizing the number of instruments simultaneously. 
\citet{hall2007information} 
proposed an efficiency and non–redundancy based 
criterion to avoid inclusion of redundant moment conditions \citep{breusch1999redundancy} which is a weakness of the orthogonality condition based criterion.
For comprehensive reviews, we suggest readers refer to \citep[Chapter 7]{hall2005generalized} and \citep{Hall15Econometricians}.

Recently the CMR models have become increasingly popular in the machine learning community \citep{Muandet20Kernel, Dikkala20:Minimax, Bennett19:DeepGMM}.
This opens up a new possibility to resolve the selection problem with modern tools in machine learning. 
Popular works includes DeepIV \citep{Hartford17:DIV}, KernelIV \citep{singh2019kernel}, DualIV \citep{Muandet20:DualIV}, and adversarial structured equation models (SEM) \citep{Liao20:NeuralSEM} in the sub-area of causal machine learning; see, also, \citet{liao2021instrumental} and references therein for related works from the reinforcement learning (RL) perspective. 
In this line of work, no instruments are exploited and the focus is mainly on the  estimation of the conditional density or solving the saddle-point reformulation of the CMR. 
On the other hand, the second line of work, which is more related to ours, includes adversarial GMM \citep{Lewis18:AGMM,Dikkala20:Minimax,zhang2020maximum,Liu18:Infinite,feng2019kernel,Uehara2020:Minimax,Kallus20:Causal} with an adversarial instrument from a function space, and DeepGMM \citep{Bennett19:DeepGMM} with fixed instruments. 
All of these methods employ flexible instruments such as neural networks \citep{Lewis18:AGMM,Dikkala20:Minimax} or RKHSs \citep{Dikkala20:Minimax,zhang2020maximum,Liu18:Infinite,feng2019kernel,Uehara2020:Minimax,Kallus20:Causal}, and the adversarial GMM selects appropriate instruments by maximizing the GMM objective function in order to obtain robust estimators. 
To the best of our knowledge, none of these works addressed the moment selection problem directly for the IV estimator.

Lastly, the KMMR objective is also related to the maximum mean discrepancy (MMD) \citep{borgwardt2006integrating} and the kernel Stein discrepancy (KSD) \citep{liu2016kernelized}, as pointed out by \citet{Muandet20Kernel}. 
The kernel selection problem for MMD was previously studied on the two-sample testing problem \citep{Gretton12Optimal}. 
The principle is to maximize the test power and later widely applied to kernel selection for different hypothesis testing, such as independence testing based on finite set independence
criterion \citep{jitkrittum2017adaptive} and goodness-of-fit testing based on finite set KSD \citep{Jitkrittum17:Linear}. 
While this method can be applied to the KMMR based hypothesis testing \citep{Muandet20Kernel}, it is not applicable in our case as we focus on parameter estimation problem.

\section{Experiments}

We demonstrate the effectiveness of our selection criterion on the IV regression task with two baselines: (1) the median heuristic employed by \citet{zhang2020maximum} and (2) Silverman's rule-of-thumb \citep{silverman1986density}. These baselines are widely used in statistics and related fields \citep{hall1991optimal}.

\subsection{Simple linear IV regression}\label{sec:exp_select_LIIS}
We first demonstrate the ability of our method to select the LIIS among a set of candidate instrument spaces. 
We consider simple linear (in parameters) IV regression models, because it is easy to analyze the global identifiability of instrument spaces.
\citet[section 2.1]{hall2005generalized} provides its details.

\textbf{Data.} We consider data generation models employed by \citet{Bennett19:DeepGMM} and adapt them to our experiments: $ Y = f^*(X)+ e+\delta$, $ X = Z+e+\gamma$,
where $Z \in \mathrm{Uniform}[-3,3]$, $e \sim \mathcal{N}(0,1)$ and  $\delta,\gamma \sim \mathcal{N}(0,0.1^2)$. The variable $e$ is the confounding variable that creates the correlation between $X$ and the residual $(Y-f^*(X))$. $Z$ is an instrumental variable correlated to $X$ and we employ two choices of function $f^*$ to enrich the test scenarios:
\begin{enumerate*}[label=(\roman*),noitemsep]
\item \texttt{linear}: $f^*(x)=x$;
   \item \texttt{quadratic}: $f^*(x)= x^2+x$.
\end{enumerate*}
Besides, we generate $n \in [100, 500, 1000]$ data for training, validation and test sets respectively. We give more experiment details in Appendix \ref{sec:exp_setting}.

\textbf{Algorithms.} We use a linear combination of polynomial basis functions to estimate the unknown true function $f^*(x) $, namely, $f_m(x) \coloneqq \sum_{i=0}^{m} c_i x^{i}$, where $c_i \in \pp{R}$ denote the unknown parameters and we consider two polynomial degrees $m \in [2,4]$. For the KMMR framework, we employ a wide range of kernels to construct instrument spaces: (i) the linear kernel (denoted as \textit{L}) $k_L(z,z') = z z'$; (ii) the polynomial kernel $k_{Pd}(z,z') = (zz'+p)^{d}$, where $p \geq 0$ is a kernel parameter, $d \in \mathbb{N}$ controls the polynomial degree and we consider $d\in[2,4]$ (denoted as \textit{P2}, \textit{P4}) and $p \in [1,2]$; (iii) the Gaussian kernel (denoted as \textit{G}) $k_G(z,z') = \exp( -(z-z')^2/(2p^2) )$ where we consider the bandwidth $p\in [0.1, 0.2, 0.5,1,2]$. LIIS is identified with this setting (see details in Appendix \ref{sec:exp_LIIS}): P2 with $p=1$ for $f_2$ and P4 with $p=1$ for $f_4$.


\textbf{Results.} We present the results on the linear $f^*$ in Fig. \ref{fig:experiment1_linear} and the quadratic  $f^*$ in Fig. \ref{fig:experiment1_quad} in Appendix. We can see that unidentifiable instrument spaces for $f_2$: L, and for $f_4$: L, P2-1, P2-2, are always tested to be unidentifiable, while other identifiable spaces have increasing significance of identifiability as data size. For $f_2$, the LIIS P2-1 is chosen as the LIIS by our method on the large datasets while due to more parameters contained in $f_4$, the LIIS P4-1 requires more data to be tested as an identifiable space significantly. This presents a difficulty of our method to identify the LIIS when the model has many parameters and the dataset is small.

\begin{figure}[t]
    \hspace{1.8cm}
    \begin{subfigure}{0.4\textwidth}
        \raggedright
        \resizebox{!}{1.5in}{
        \begin{tikzpicture}[trim left=0cm,trim right=0cm]
            \begin{axis}[%
                width=2.1in, height=1.6in,
                grid=major,
                scatter/classes={%
                K100={mark=*,draw=black,scale=1.4, fill=blue},
                K500={mark=triangle*,draw=black,scale=1.8,
                fill=green!60!black}, 
                K1000={mark=square*,draw=black,scale=1.2, fill=red}
                },
                    ymode=log,
                    ymax=1.1, ymin=1e-9,
                    xmax=9,xmin=0,
                    xtick=data,
                    xticklabels={L,P2-1, P2-2,P4-1, P4-2,  G-2, G-1, G-0.5, G-0.2, G-0.1},
                    xticklabel style={rotate=90,anchor=east,
                    font=\scriptsize},
                    yticklabel style={
                    font=\scriptsize},
                    ylabel = {Normalized ITC},
                    ylabel near ticks,
                    legend pos= south east,
                    legend style={nodes={scale=0.7, transform shape}},
                    label style={
                    font=\footnotesize}
                ]
            \addplot[scatter,only marks,%
                scatter src=explicit symbolic]
                table[x=x,y=y,meta=label]
                {linear_itc_s_f2.dat};
                \addplot[domain=0:20,
                    samples=10, color=red,-,dashed,line width=2pt]
                    {7.477043090647247e-05};
            \node[right,blue] at (axis cs: 5,0.0005553492777589242) {\textbf{*}};
            \node[right,green!60!black] at (axis cs: 1,0.0005551533141673801) {\textbf{*}};
            \node[right,red] at (axis cs: 1,0.003228698529923325) {\textbf{*}};
             \legend{n=100,n=500,n=1000};
            \end{axis}
            \end{tikzpicture}}
        \end{subfigure}
        \begin{subfigure}{0.4\textwidth}
            \raggedright
            \resizebox{!}{1.5in}{
            \begin{tikzpicture}[trim left=0cm,trim right=0cm]
                \begin{axis}[%
                    width=2.1in, height=1.6in,
                    grid=major,
                    scatter/classes={%
                    K100={mark=*,draw=black,scale=1.4, fill=blue},
                    K500={mark=triangle*,draw=black,scale=1.8,
                    fill=green!60!black}, 
                    K1000={mark=square*,draw=black,scale=1.2, fill=red}
                    },
                        ymode=log,
                        ymax=1.1, ymin=1e-7,
                        xmax=9,xmin=0,
                        xtick=data,
                        xticklabels={L,P2-1, P2-2,P4-1, P4-2,  G-2, G-1, G-0.5, G-0.2, G-0.1},
                        xticklabel style={rotate=90,anchor=east,
                        font=\scriptsize},
                        yticklabel style={
                        font=\scriptsize},
                        ylabel near ticks,
                    ]
                \addplot[scatter,only marks,%
                    scatter src=explicit symbolic]
                    table[x=x,y=y,meta=label]
                    {linear_itc_s_f4.dat};
                    \addplot[domain=0:20,
                        samples=10, color=red,-,dashed,line width=2pt]
                        {0.013227436169095323};
                \node[right,blue] at (axis cs: 7,0.07258992895562752) {\textbf{*}};
                \node[right,green!60!black] at (axis cs: 7,0.15463375784838817) {\textbf{*}};
                \node[right,red] at (axis cs: 6,0.06506980344986206) {\textbf{*}};
                \end{axis}
                \end{tikzpicture}}
            \end{subfigure}
            \caption{ITC evaluated on the linear function $f^*$. The left and right plots employ $f_2$ and $f_4$ in the estimation respectively. We use [\textit{kernel}]-[\textit{parameter}] to denote different kernels and normalize all values in each plot to $[0,1]$. The symbols (*) on the \textit{right} of nodes denote the selected LIISs. The red dash lines denote the quantile corresponding to the significance level $\alpha=0.05$.} 
            \label{fig:experiment1_linear}
\end{figure}
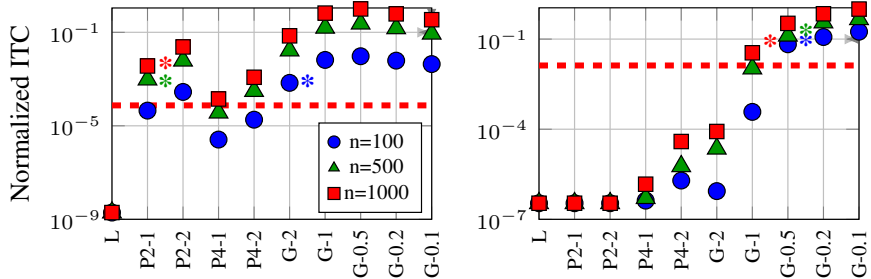

\begin{table}[t!]
    \centering
        \caption{The mean square error (MSE) $\pm$ one standard deviation ($n=500$, $f_4$).}
        \label{tab:result_low_dim_large}
        \resizebox{0.95\textwidth}{!}{
        \begin{tabular}{l l c c c c} 
        \toprule
        \multirow{2}{*}{\textbf{Scenario}} &\multirow{2}{*}{\textbf{Algorithm}} & \multicolumn{4}{c}{\textbf{True Function $f^*$}} \\
        & & abs & linear & quad &sin \\ 
        \midrule
        \multirow{3}{*}{LS} &\texttt{Silverman}    & 0.023 $\pm$ 0.006 & 0.006 $\pm$ 0.005 & 0.006 $\pm$ 0.007 & 0.032 $\pm$ 0.012 \\
        &\texttt{Med-Heuristic} & 0.023 $\pm$ 0.007 & 0.006 $\pm$ 0.006 & 0.006 $\pm$ 0.008 & 0.032 $\pm$ 0.010\\
        &\texttt{Our Method} & 0.023 $\pm$ 0.007 & 0.006 $\pm$ 0.006 & 0.006 $\pm$ 0.008 & 0.031 $\pm$ 0.010\\
        \midrule
        \multirow{3}{*}{LW} &\texttt{Silverman}    & 0.055 $\pm$ 0.059 & 0.032 $\pm$ 0.023 & 0.017 $\pm$ 0.011 & 0.058 $\pm$ 0.050 \\
        &\texttt{Med-Heuristic} & 0.214 $\pm$ 0.019 & 0.066 $\pm$ 0.011 & 0.037 $\pm$ 0.018 & 0.101 $\pm$ 0.012 \\
        &\texttt{Our Method} & \textbf{0.024} $\pm$ \textbf{0.010} & \textbf{0.015} $\pm$ \textbf{0.016} & \textbf{0.009} $\pm$ \textbf{0.010} & \textbf{0.019} $\pm$ \textbf{0.022} \\
        \midrule
        \multirow{3}{*}{NS} &\texttt{Silverman}    & 7.384 $\pm$ 18.271 & 0.137 $\pm$ 0.118 & 0.595 $\pm$ 0.733 & 0.539 $\pm$ 0.913 \\
        &\texttt{Med-Heuristic} & 0.070 $\pm$ 0.044 & 0.021 $\pm$ 0.019 & 0.019 $\pm$ 0.011 & 0.074 $\pm$ 0.053 \\
        &\texttt{Our Method} & \textbf{0.039} $\pm$ \textbf{0.019} & \textbf{0.006} $\pm$ \textbf{0.004} & \textbf{0.007} $\pm$ \textbf{0.003} & \textbf{0.028} $\pm$ \textbf{0.027} \\
        \bottomrule
        \end{tabular}}
        \label{tab:mse_f4_500}
    \end{table}

\subsection{Robustness of parameter estimation with linear models}
We compare our method and baselines on parameter estimation with the linear IV model.

\textbf{Settings.}  We employ the afore-defined linear model $f_4$ for IV regression and measure the mean square error (MSE) between the optimized $f_4$ and the ground truth $f^*$. We adapt the afore data generation models as below:
    $Y = f^*(X)+ e+\delta$, with $ X = d^{-1}\sum_{i=1}^{d}g(Z_i)+e+\gamma$,
where $Z:=(Z_1,\cdots,Z_d) \sim \mathrm{Uniform}([-3,3]^d)$, and other variables have the same definitions as the last subsection. We consider two more choices of $f^*$:
(iii) \texttt{abs}: $f^*(x)= |x|$, and
(iv) \texttt{sin}: $f^*(x)=\sin(x)$,
apart from (i) \texttt{linear} and (ii) \texttt{quadratic}  $f^*$ in the last subsection. We sample $n\in [100,500]$ data for training, validation and test sets respectively.
We design three evaluation scenarios: (a) $d=1$, $g(Z)=Z$, (b) $d=6$ , $g(Z)=Z$, (c) $d=1$ , $g(Z)=\sin(Z)$. In (a), $Z$ is a strong IV linearly correlated with $X$, which is an ideal data generation process, and we refer to this scenario as the linearly strong (LS) IV scenario. The scenario (b) has $d=6$ weak IVs linearly correlated with $X$ (referred to as a linearly weak IV scenario, shortened as LW) and it is common in the real-world applications, such as the genetic variants employed in Mendelian randomization which are known as weak IVs \citep{Kuang20:Ivy,Hartford20:WeakIV,Burgess20:GIV}. The scenario (c) takes the nonlinearly strong correlation between $Z$ and $X$ in account (referred to as nonlinearly strong IV scenario, shortened as NS) and this consideration is also common, e.g., in the two-step least square methods, nonlinear models are often employed to fit the relation between $Z$ and $X$ \citep{Hartford17:DIV,singh2019kernel}.

\textbf{Results.} Results are shown in Table \ref{tab:mse_f4_500} ($n=500$) and \ref{tab:mse_f4_100} ($n=100$, in Appendix). 
Two baselines and our method have very close performance in the LS scenario. 
In contrast, weak IVs in the LW scenario have a significantly negative effect on the median-heuristic method, where our method performs stably and Silverman's rule of thumb a bit less stably than our method. Moreover, the Silverman's rule of thumb suffers from the nonlinear correlation between $Z$ and $X$ in the NS scenario and the median heuristic method performs better in this scenario than in the LW scenario. Our method still performs stably and well.
The experiment results show robustness of our method on parameter estimation in the different scenarios and the superiority of our method which can select any kernel flexibly, whereas the baselines can only handle the Gaussian kernels in principle. Therefore, our method is preferable for the task of parameter estimation than the two baselines. 


\subsection{Robustness of parameter estimation with neural networks}
    We further assess the effectiveness of our method on parameter estimation with the neural network (NN), which  is a commonly-used model in machine learning.
    Due to complicated structures of NNs, the problem is harder than the previous one.

     \textbf{Settings.} We employ the weak IV scenario with $d=2$ from the last subsection and generates $n=500$ samples for training, validation and test datasets respectively. For a neat demonstration and to avoid e.g. training difficulties with deep NN models, we employ two relatively simple models: (i) one has just one hidden layer with 10 latent units, denoted as $N_{10}$, and (ii) the other has two hidden layers with 5 latent units for each layer, denoted as $N_{55}$. The sigmoid activation function is used for hidden layers. Both models are sufficient to fit the simple true causal functions $f^*(X)$. We find that there are often no instrument spaces being identifiable significantly due to many parameters. Therefore, we minimize the ratio $\mathrm{KEIC}/\mathrm{ITC}$ defined in section \ref{sec:selection-criteria} to select the LIIS.
     
     \textbf{An approximation to ITC.} We approximate ITC to reduce computational burden due to many parameters in  NNs. 
     We consider a common form of NNs $f(x) = W_0 \Phi(x)+b_0$ where $\Phi(x) = \sigma_h(b_h+W_h \sigma_{h-1}(\cdots \sigma_1 (b_1 + W_1x))) $ denotes a depth-$h$ structure with weights $\bm W$, biases $\bm b$ and activation functions $\bm \sigma$. Therefore, we view $\Phi(x)$ as basis functions, whose parameters are updated during the training process while fixed in computing the ITC. As a result, we only use the gradients on $W_0$ and $b_0$ to evaluate the ITC, which is an approximation to the ITC computed on gradients on all parameters. This can accelerate the computation of the ITC for a NN and we assess the performance of parameter estimation with this approximation.

    \textbf{Results.} The MSEs between the optimized NN $\hat{f}_{\mathrm{NN}}$ and $f^*$ are reported in Table \ref{tab:mse_NN}. First, we find that minimizing the ratio of $\mathrm{KEIC}/\mathrm{ITC}$ indeed helps to reduce the biases of parameter estimation with NNs. Compared with two baselines, our method performs stably on different datasets and on different NN structures. This provides further evidence that our method improve the practical performance of the KMMR. Second, our approximate ITC can work for the employed NNs.

    \begin{table}[t]
        \centering
            \caption{The mean square error (MSE) $\pm$ one standard deviation ($n=500$, NN). }
            \label{tab:result_low_dim_large}
            \resizebox{0.95\textwidth}{!}{
            \begin{tabular}{l l c c c c} 
            \toprule
            \multirow{2}{*}{NN} &\multirow{2}{*}{\textbf{Algorithm}} & \multicolumn{4}{c}{\textbf{True Function $f^*$}} \\
            & & abs & linear & quad &sin \\ 
            \midrule
            \multirow{3}{*}{$\mathrm{NN}_{10}$} &\texttt{Silverman}    & 0.328$\pm$0.152 & 0.166$\pm$0.107 & 0.031$\pm$0.020 & 0.327$\pm$0.098 \\
            &\texttt{Med-Heuristic} & 0.231$\pm$0.030 & 0.045$\pm$0.021 & 0.012$\pm$0.004 & 0.179$\pm$0.034 \\
            &\texttt{Our Method} & \textbf{0.041}$\pm$\textbf{0.027} & \textbf{0.027}$\pm$\textbf{0.016} & \textbf{0.006}$\pm$\textbf{0.003} & \textbf{0.058}$\pm$\textbf{0.045} \\
            \midrule
            \multirow{3}{*}{$\mathrm{NN}_{55}$} &\texttt{Silverman}    & 0.444$\pm$0.255 & 0.102$\pm$0.074 & 0.037$\pm$0.018 & 0.630$\pm$0.398 \\
            &\texttt{Med-Heuristic} & 0.187$\pm$0.044 & 0.039$\pm$0.018 & 0.013$\pm$0.004 & 0.145$\pm$0.039 \\
            &\texttt{Our Method} & \textbf{0.036}$\pm$\textbf{0.013} & \textbf{0.013}$\pm$\textbf{0.003} & \textbf{0.007}$\pm$\textbf{0.003} & \textbf{0.035}$\pm$\textbf{0.024}\\
            \bottomrule
            \end{tabular}}
            \label{tab:mse_NN}
        \end{table}
\section{Conclusions}
The conditional moment restriction (CMR) is ubiquitous in many fields and the kernel maximum moment restriction (KMMR) is a promising framework to deal with the CMR in its easy-to-use-and-analyze form and practical outperformance. However, the optimal choice of the instrument space is a challenge and affects the effectiveness of the framework. The present work propose a systematic procedure to select the instrument space for instrumental variable (IV) estimators. We first define a selection principle of the \textit{least identifiable instrument space} (LIIS) that identifies the estimated model parameters with the least space complexity. To determine the LIIS among a set of candidates, we propose the least identification selection criterion (LISC) as a combination of two criteria: (i) the identification test criterion (ITC) to check identifiability; (ii) the kernel effective information criterion (KEIC) consisting of the effective dimension and the risk function value to check the CMR condition and the effective dimension. The consistency of our method is explored and the experiments provide evidence for the effectiveness of our method on identifying the LIIS and on the parameter estimation for IV estimators.

\bibliographystyle{abbrvnat}
\bibliography{references}
\section*{Checklist}

\begin{enumerate}
    \item For all authors...
    \begin{enumerate}
      \item Do the main claims made in the abstract and introduction accurately reflect the paper's contributions and scope?
        \answerYes{As mentioned in the abstract and introduction, this paper is to address the kernel instrument space selection problem. The problem arises recently from the kernel maximum moment restriction framework for instrumental variable estimators and is different from traditional selection problem on instruments.}
      \item Did you describe the limitations of your work?
        \answerYes{In the experiment section 6.1, we present that our identification test criterion can't identify the optimal instrument space for models with many parameters  given small datasets. }
      \item Did you discuss any potential negative societal impacts of your work?
        \answerNA{}
      \item Have you read the ethics review guidelines and ensured that your paper conforms to them?
        \answerYes{}
    \end{enumerate}
    
    \item If you are including theoretical results...
    \begin{enumerate}
      \item Did you state the full set of assumptions of all theoretical results?
        \answerYes{For our theoretical results, we define a series of assumptions such as \ref{asmp:global_ident} and \ref{asmp:identifiable}, and include extra conditions in the statement of the theorems.}
        \item Did you include complete proofs of all theoretical results?
        \answerYes{We put proofs into the appendix.}
    \end{enumerate}
    
    \item If you ran experiments...
    \begin{enumerate}
      \item Did you include the code, data, and instructions needed to reproduce the main experimental results (either in the supplemental material or as a URL)?
        \answerYes{We put codes for our experiments into the supplemental material.}
      \item Did you specify all the training details (e.g., data splits, hyperparameters, how they were chosen)?
        \answerYes{All training details are included in the experiment section and additional sections in the appendix.}
        \item Did you report error bars (e.g., with respect to the random seed after running experiments multiple times)?
        \answerYes{We report error bars and provide the random seed and the number of repeated experiments.}
        \item Did you include the total amount of compute and the type of resources used (e.g., type of GPUs, internal cluster, or cloud provider)?
        \answerNo{Our experiment results don't rely on the type of computational resources.}
    \end{enumerate}
    
    \item If you are using existing assets (e.g., code, data, models) or curating/releasing new assets...
    \begin{enumerate}
      \item If your work uses existing assets, did you cite the creators?
        \answerYes{}
      \item Did you mention the license of the assets?
        \answerNA{We implement baselines and our method ourselves, and our data are simulated.}
      \item Did you include any new assets either in the supplemental material or as a URL?
        \answerYes{We put our code for experiments in the supplemental material.}
      \item Did you discuss whether and how consent was obtained from people whose data you're using/curating?
        \answerNA{See the answer for 4(b).}
      \item Did you discuss whether the data you are using/curating contains personally identifiable information or offensive content?
        \answerNA{Our data are simulated without any personally identifiable information or offensive content.}
    \end{enumerate}
    
    \item If you used crowdsourcing or conducted research with human subjects...
    \begin{enumerate}
      \item Did you include the full text of instructions given to participants and screenshots, if applicable?
        \answerNA{}
      \item Did you describe any potential participant risks, with links to Institutional Review Board (IRB) approvals, if applicable?
        \answerNA{}
      \item Did you include the estimated hourly wage paid to participants and the total amount spent on participant compensation?
        \answerNA{}
    \end{enumerate}    
\end{enumerate}
\appendix
\onecolumn

\doparttoc \faketableofcontents
\addcontentsline{toc}{section}{Appendix} 
\part{Appendix  \\ {\large Instrument Space Selection for Kernel Maximum Moment Restriction}} 
\parttoc 

\section{Derivation Process of ITC} \label{app:itc}

To derive an asymptotic distribution of $\hat{T}$, we need an asymptotic distribution of $\hat{F}_\hbsph(\hat{\theta})$, shown as follows.
As preparation, let $\mathrm{vec}:\pp{R}^{s\times t} \to \pp{R}^{st \times 1} $ be a vectorization operator by a rule of $[\mathrm{vec}(A)]_{(i-1)t+j,1} = A_{i,j}$, $i=1,\cdots,s$, $j=1,\cdots,t$.
\begin{lemma}\label{lemma:asym_dist_F}
    For  $\hat{\theta} \rightarrow \theta$ almost surely and a non-zero matrix $\Omega = \ep_{S} \left[ \mathrm{vec}(u_{\theta}(S))\mathrm{vec}(u_{\theta}(S))^{\top}\right]-\ep_{S} \left[ \mathrm{vec}(u_{\theta}(S))\right]\ep_{S} \left[ \mathrm{vec}(u_{\theta}(S))\right]^{\top} \in \mathbb{R}^{c^2 \times c^2}$, we have $\sqrt{n} \mathrm{vec}(\hat{F}_\hbsph(\hat{\theta}) - {F}_\hbsph({\theta})) \dto \mathcal{N}(0, \Omega)$.
\end{lemma}

The asymptotic variance $\Omega$ is an essential term to determine an asymptotic distribution of $\hat{T}$. We need an assumption on $\Omega$ under the non-full-rank hypothesis $H_0$. Let $C$ be the eigenvector of $F_\hbsph(\theta)$ corresponding to the eigenvalue $\lambda_c=0$ under $H_0$. We make the following rank assumption.
\begin{assumption}[rank condition] \label{asmp:non_degenerate_Omega}
    Under the null hypothesis $H_0$, the matrix $(C \otimes C)^{\top} \Omega (C \otimes C)$ is non-zero, where $\otimes$ is the Kronecker product.
\end{assumption}
An equivalent statement of Assumption \ref{asmp:non_degenerate_Omega} is that the $c^2 \times 1$ vector $(C \otimes C)$ does not lie in the space spanned by
 the eigenvectors associated with the zero eigenvalue of $\Omega$.  Therefore, if $\Omega$ is positive definite, Assumption \ref{asmp:non_degenerate_Omega} is automatically satisfied. This assumption is required to obtain a non-degenerate asymptotic distribution of the test statistic. Let $\hat{C}$ be the eigenvalue of $\hat{F}_{\hbsph}(\hat{\theta})$ corresponding to the eigenvalue $\hat{\lambda}_c\geq 0$. Under the hypothesis $H_0$ and the rank assumption \ref{asmp:non_degenerate_Omega}, we obtain the asymptotic distribution.
 
\begin{theorem}(asymptotic distribution of $\hat{T}$)\label{thm:rank-asymptotic}
    Suppose conditions of Lemma \ref{lemma:asym_dist_F}, assumption \ref{asmp:non_degenerate_Omega} and the null-hypothesis $H_0$ hold. Then we obtain the asymptotic distribution: $n \hat{T} \dto \Lambda Z^2$ where $\Lambda\coloneqq (C \otimes C)^\top \Omega (C \otimes C)$ and
     $Z$ is a  standard normal random variable.
\end{theorem}

By the asymptotic distribution, we can handle the type-I error.
For a significance level $\alpha \in (0,1)$, let $Q_{1-\alpha}$ be a $(1-\alpha)$-quantile of $\Lambda Z^2$.
Then, the following result shows that, the type-I error is asymptotically equal to the arbitrary value $\alpha \in [0,1]$. 
\begin{theorem} \label{thm:valid_test}
    Suppose Assumption \ref{asmp:non_degenerate_Omega} holds.
    Let $\mathbb{P}_0$ be a probability measure of data under the null-hypothesis $H_0$.
    Then, for any $\alpha \in (0,1)$, we obtain $\mathbb{P}_0 (n \hat{T} \geq Q_{1-\alpha} ) \to  \alpha$
    as $n \to \infty$.
\end{theorem}

\section{Proofs}

\subsection{Proof of Lemma \ref{lemma:asym_dist_F}}

\begin{proof}
We first decompose the objective into two terms (i) and (ii):
\begin{align}
    \sqrt{n} \mathrm{vec}(\hat{F}_\hbsph(\hat{\theta}) - {F}_\hbsph({\theta})) = \underbrace{\sqrt{n} \mathrm{vec}(\hat{F}_\hbsph(\hat{\theta}) -F_\hbsph(\hat{\theta}))}_{\text{(i)}}+\underbrace{\sqrt{n}\mathrm{vec}(F_\hbsph(\hat{\theta}) - {F}_\hbsph({\theta}))}_{\text{(ii)}}
\end{align}
The term (ii) converges to zero as $n \to \infty$ given $\hat{\theta} \rightarrow \theta$ almost surely. By Slutsky's theorem \citep[Lemma 2.8]{Vaart00:Asymptotic}, the asymptotic distribution depends on the term (i). Let $\ep_n$ denote the empirical expectation and the asymptotic distribution of (i) is obtained by the central limit theorem
\begin{align}
&\sqrt{n} \mathrm{vec}(\hat{F}_{\hbsph}(\hat{\theta})-F_{\hbsph}(\hat{\theta})) 
=\sqrt{n} \left(\ep_{n}\left[\mathrm{vec}( u_{\hat{\theta}}(S)) \right]-\ep_{S}\left[\mathrm{vec}( u_{\hat{\theta}}(S))\right]\right)\dto \mathcal{N}(\bm{0}, \Omega),
\\
&
\Omega=\ep_{S} \left[ \mathrm{vec}(u_{\theta}(S)) \mathrm{vec}(u_{\theta}(S))^{\top}\right]-\ep_{S} \left[ \mathrm{vec}( u_{\theta}(S))\right]\ep_{S} \left[\mathrm{vec}( u_{\theta}(S))\right]^{\top}.
\end{align}
\end{proof}

\subsection{Proof of Theorem \ref{thm:rank-asymptotic}}

\begin{proof}
    We rewrite $\hat{T}^{1/2}\coloneqq \hat{\lambda} = \hat{C} \hat{F}_{\hbsph}(\hat{\theta})\hat{C}$ and $T^{1/2} \coloneqq \lambda = C F_{\hbsph}(\theta)C$. We  further rewrite $\sqrt{n}(\hat{T}-T)$ as below
    \begin{align}
        \sqrt{n}\left(\hat{T}^{1/2}-T^{1/2} \right)&=\sqrt{n}(\hat{C} \hat{F}_{\hat{\hbsph}}(\theta)\hat{C} - C F_{\hbsph}(\theta)C)
        \\
        &=\underbrace{\sqrt{n}(\hat{C} \hat{F}_{\hbsph}(\hat{\theta})\hat{C} - C \hat{F}_{\hbsph}(\hat{\theta})C)}_{\text{(i)}} + \underbrace{\sqrt{n}(C \hat{F}_{\hbsph}(\hat{\theta})C-C F_{\hbsph}(\theta)C )}_{\text{(ii)}}.
    \end{align}
    The term (i) converges to $0$ as $n \to \infty$ because $C \hat{F}_{\hbsph}(\hat{\theta})C\to C F_{\hbsph}(\theta)C = \lambda_c$ and $\hat{C} \hat{F}_{\hbsph}(\hat{\theta})\hat{C}=\hat{\lambda}_c \to \lambda_c$ based on the consistency of $\hat{F}_{\hbsph}(\hat{\theta})$ to $F_{\hbsph}(\theta)$ given in Lemma \ref{lemma:asym_dist_F}. The term (ii) has an asymptotic distribution following by Lemma \ref{lemma:asym_dist_F} and Assumption \ref{asmp:non_degenerate_Omega}:
    \begin{align}
        \sqrt{n}\left(C \hat{F}_{\hbsph}(\hat{\theta})C-C F_{\hbsph}(\theta)C \right) &= \sqrt{n}(C\otimes C)^{\top} \mathrm{vec}(\hat{F}_{\hbsph}(\hat{\theta})-F_{\hbsph}(\theta))
        \\
        &\dto \mathcal{N}(0, (C\otimes C)^{\top} \Omega (C\otimes C))
        \\
        &\quad = \left[(C\otimes C)^{\top} \Omega (C\otimes C)\right]^{1/2}\mathcal{N}(0, 1).
    \end{align}
    Because the null-hypothesis holds, i.e., the matrix $F_{\hbsph}(\theta)$ is non-full-rank, we obtain that $CF_{\hbsph}(\theta)C=\lambda_c=0$. Therefore, $\sqrt{n}\hat{T} \dto \Lambda^{1/2} Z$ by Slutsky's theorem and $n\hat{T} \dto \Lambda Z^2$ by \citep[Lemma 3.2]{Vuong1989Likelihood}, where $\Lambda = (C\otimes C)^{\top} \Omega (C\otimes C)$ and $Z$ is a standard normal random variable.
\end{proof}

\subsection{Proof of Theorem \ref{thm:valid_test}}

\begin{proof}
The conclusion follows directly from the asymptotic distribution of $n\hat{T}$.
\end{proof}

\subsection{Proof of Theorem \ref{thm:consistency_ITC}}

\begin{proof}
    By Assumption \ref{asmp:global_ident} and consistency of $R_{\hbsph}(\theta)$ to $\mathrm{CMR}(\theta)$, we know that $R_{\hbsph}(\theta)$ has a unique minimizer at $\theta_0$ which satisfies $\mathrm{CMR}(\theta_0)=0$. Further since $\Theta$ is compact, $R_{n,\hbsph}(\theta)$ converges to $R_{\hbsph}(\theta)$ uniformly in probability and $R_{\hbsph}(\theta)$ is continuous (due to $R_{\hbsph}(\theta)$ is finite and $\varphi_{\theta}$ is differentiable implicitly assumed), we obtain that the empirical $\hat{\theta} \pto \theta_0$ by Theorem 2.1 in \citet{Newey94:Large}. Therefore, as $n\to \infty$, $\mathrm{rank}(\hat{F}_{\hbsph}(\hat{\theta})) \pto \mathrm{rank}(F_{\hbsph}(\theta_0)))$. Therefore, the ITC consistently estimates the identifiability in probability. 
\end{proof}

\subsection{Proof of Theorem \ref{thm:consistency_E}}

\begin{proof}
By the existing result \citep[Theorem 3.4]{baker1977numerical}  that $\lambda_{i}^{\mathrm{mat}}/n \to \lambda_i^{K}$ as $n \to \infty$ where $\lambda_i^{\mathrm{mat}}$ is the $i$-th ordered eigenvalue of $K_{\bm z}$, we know that $\mathrm{Tr}(K_{\bm z})/n=\sum_{i=1}^{n} \lambda_i^{\mathrm{mat}}/n$ is a consistent estimator of $\sum_{i=1}^{\infty} \lambda_i^K$. Similarly, $(\mathrm{Tr}(K_{\bm z}^2)/n^2)^{1/2}=(\sum_{i=1}^{n} (\lambda_i^{\mathrm{mat}})^2/n^2)^{1/2}$ consistently estimates $(\sum_{i=1}^{\infty} (\lambda_i^K)^2)^{1/2}$. Therefore, $\mathrm{Tr}(K_{\bm z}){\mathrm{Tr}(K_{\bm z}^2)^{-1/2}}$ is a consistent estimator of $E_k$.
\end{proof}

\subsection{Proof of Theorem \ref{thm:keic}}

\begin{proof}
We define $j^* \in \mathcal{M}$ as an optimal index of the optimal RKHS instrument space, namely, $\hbsph_{j^*} = \hbsph^l$. With the above settings, the statement of Theorem \ref{thm:keic} is equivalent to show $\mathrm{P}(\mathrm{KEIC}(\hbsph_j) > \mathrm{KEIC}(\hbsph_{j^*})) \to 1$ as $n \to \infty$, for any $j \in \mathcal{M}$ such that $j \neq j^*$.
We define $E_j$ is an effective dimension of $\hbsph_j$ and $\hat{E}_{j} := \mathrm{Tr}(K_{\bm z}){\mathrm{Tr}(K_{\bm z}^2)^{-1/2}}$ with a kernel $k$ corresponding to the RKHS $\hbsph_j$.
For $j\neq j^*$ and $j^*$, we study the following difference
\begin{align*}
    \mathrm{KEIC}(\hbsph_j) - \mathrm{KEIC}(\hbsph_{{j^*}}) 
    = n \underbrace{(\hat{R}_{\hbsph_j}(\hat{\theta})-\hat{R}_{\hbsph_{{j^*}}} (\hat{\theta}))}
    _{\coloneqq \Delta_R (j,j^*)}  
    + \underbrace{(\hat{E}_j - \hat{E}_{j^*})}_{\coloneqq \Delta_E (j,j^*)} \log n.
\end{align*}
Given the condition that an unique least identifiable instrument space exists and the consistency of $\hat{E}$ to $E$ shown in Theorem \ref{thm:consistency_E}, we obtain that $\mathrm{P}(\hat{E}_j-\hat{E}_{j^*} > 0) \to 1$ and $\Delta_E (j,j^*)$ converges to some constant. Hence, it follows that $\Delta_E (j,j^*) \log n := \Omega(\log n)$ with $\Omega(\cdot)$ in complexity theory. By the assumption \ref{asmp:global_ident} that $\mathrm{CMR}(\theta)=0$ has an unique solution say $\theta_0$ and $H_j$ is identifiable, the minimizer $\hat{\theta} = \argmin_{\theta} \hat{R}_{\hbsph_j}(\theta)$ converges to $\theta_0$. As a result, $n \Delta_R(j,j^*) = O(1)$ by the asymptotic distribution in the conclusion (2) of Theorem 4.1 in \citet{Muandet20Kernel}.
 Hence the term $\Delta_E (j,j^*) \log n$ has a larger order than $n \Delta_R(j,j^*)$ and we obtain that $\mathrm{P}(\mathrm{KEIC}(\hbsph_j) > \mathrm{KEIC}(\hbsph_{j^*})) \to 1$ as $n \to \infty$ for $j \neq j^*$.


\end{proof}

\begin{figure}[t]
    \hspace{1.8cm}
    \begin{subfigure}{0.4\textwidth}
        \raggedright
        \resizebox{!}{1.8in}{
        \begin{tikzpicture}[trim left=0cm,trim right=0cm]
            \begin{axis}[%
                width=2.1in, height=1.6in,
                grid=major,
                scatter/classes={%
                K100={mark=*,draw=black,scale=1.4, fill=blue},
                K500={mark=triangle*,draw=black,scale=1.8,
                fill=green!60!black}, 
                K1000={mark=square*,draw=black,scale=1.2, fill=red}
                },
                    ymode=log,
                    ymax=1.1, ymin=1e-9,
                    xmax=9,xmin=0,
                    xtick=data,
                    xticklabels={L,P2-1, P2-2,P4-1, P4-2,  G-2, G-1, G-0.5, G-0.2, G-0.1},
                    xticklabel style={rotate=90,anchor=east,
                    font=\scriptsize},
                    yticklabel style={
                    font=\scriptsize},
                    ylabel = {Normalized ITC},
                    ylabel near ticks,
                    legend pos= south east,
                    legend style={nodes={scale=0.7, transform shape}},
                ]
            \addplot[scatter,only marks,%
                scatter src=explicit symbolic]
                table[x=x,y=y,meta=label]
                {quad_itc_s_f2.dat};
                \addplot[domain=0:20,
                    samples=10, color=red,-,dashed,line width=2pt]
                    {8.150160283636484e-05};
            \node[right,blue] at (axis cs: 5,0.0005553492777589242) {\textbf{*}};
            \node[right,green!60!black] at (axis cs: 1,0.0005551533141673801) {\textbf{*}};
            \node[right,red] at (axis cs: 1,0.003228698529923325) {\textbf{*}};
            \end{axis}
            \end{tikzpicture}}
        \end{subfigure}
        \hspace{1cm}
        \begin{subfigure}{0.4\textwidth}
            \raggedright
            \resizebox{!}{1.85in}{
            \begin{tikzpicture}[trim left=0cm,trim right=0cm]
                \begin{axis}[%
                    width=2.1in, height=1.6in,
                    grid=major,
                    scatter/classes={%
                    K100={mark=*,draw=black,scale=1.4, fill=blue},
                    K500={mark=triangle*,draw=black,scale=1.8,
                    fill=green!60!black}, 
                    K1000={mark=square*,draw=black,scale=1.2, fill=red}
                    },
                        ymode=log,
                        ymax=1.1, ymin=1e-7,
                        xmax=9,xmin=0,
                        xtick=data,
                        xticklabels={L,P2-1, P2-2,P4-1, P4-2,  G-2, G-1, G-0.5, G-0.2, G-0.1},
                        xticklabel style={rotate=90,anchor=east,
                        font=\scriptsize},
                        yticklabel style={
                        font=\scriptsize},
                        ylabel near ticks,
                        legend pos= south east,
                        legend style={nodes={scale=0.7, transform shape}},
                    ]
                \addplot[scatter,only marks,%
                    scatter src=explicit symbolic]
                    table[x=x,y=y,meta=label]
                    {quad_itc_s_f4.dat};
                    \addplot[domain=0:20,
                        samples=10, color=red,-,dashed,line width=2pt]
                        {0.020323668173503236};
                        \legend{n=100,n=500,n=1000};
                \node[right,blue] at (axis cs: 7,0.02258992895562752) {\textbf{*}};
                \node[right,green!60!black] at (axis cs: 7,0.15463375784838817) {\textbf{*}};
                \node[right,red] at (axis cs: 6,0.04506980344986206) {\textbf{*}};
                \end{axis}
                \end{tikzpicture}}
            \end{subfigure}
            \caption{ITC evaluated on the quadratic function $f^*$. The left and right plots employ $f_2$ and $f_4$ in the estimation respectively. We use [\textit{kernel}]-[\textit{parameter}] to denote different kernels and normalize all values in each plot to $[0,1]$. The symbols (*) on the \textit{right} of nodes denote the selected LIISs. The red dash lines denote the quantile corresponding to the significance level $\alpha=0.05$.} 
            \label{fig:experiment1_quad}
\end{figure}
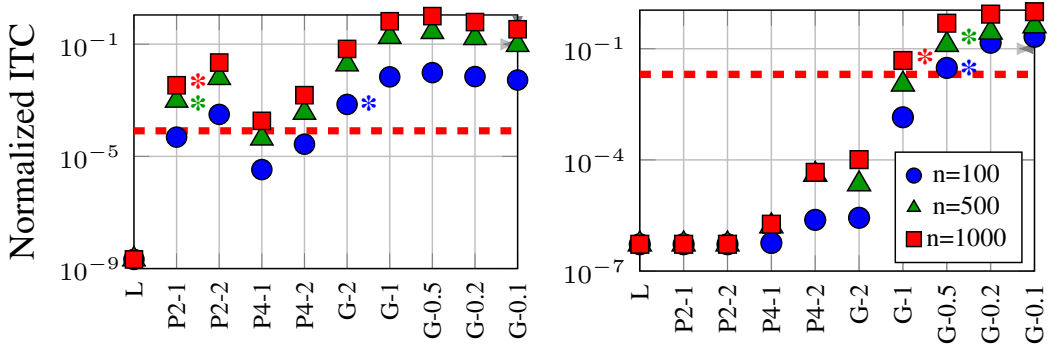

\section{More Details on Experiments}
\subsection{LIIS for Section \ref{sec:exp_select_LIIS}} \label{sec:exp_LIIS}
Under the experiment settings of Section \ref{sec:exp_select_LIIS}, an identifiable instrument space should satisfy (i) existence of column vectors of parameters $\bm{c}\coloneqq [c_i]_{i=1}^{m}$ which meet the CMR condition \eqref{eq:cmr} and (ii) uniqueness of $\bm c$, namely, $\mathrm{rank}\{\ep[ (\partial f_m(X) / \partial \bm c) (\partial f_m(X) / \partial \bm c^{\top})  k(Z,Z')] \}= m+1$ \citep[Assumption 2.3]{hall2005generalized}. 
Since the same models used for estimation and data generation, (i) immediately holds, and (ii) also holds when a number of instruments is more than that of parameters.
As the linear kernel has the feature map function $\phi_{\mathrm{L}}$ projecting $x$ to a one-dimensional space, using it for the instrument space is equivalent to employing $h(Z)=Z$ as the instrument. The condition (ii) fails due to the number of instruments is less than that of parameters of both $f_2$ and $f_4$. Moreover, using the candidate parameters, P4 identifies the parameters $\bm c$ of $f_2,\,f_4$, while P2 fails to identify those of $f_4$; G can identify parameters of $f_2,\,f_4$ using any candidate lengthscale. Considering the complexity measured by the effective dimension \eqref{def:effective_dim}, we obtain the LIISs: P2 with $p=1$ for $f_2$ and P4 with $p=1$ for $f_4$. We can approximate the effective dimension via a large sample.

\begin{table}[b!]
    \centering
        \caption{The mean square error (MSE) $\pm$ one standard deviation ($n=100$, $f_4$).}
        \label{tab:result_low_dim_large}
        \resizebox{\textwidth}{!}{
        \begin{tabular}{l l c c c c} 
        \toprule
        \multirow{2}{*}{\textbf{Scenario}} &\multirow{2}{*}{\textbf{Algorithm}} & \multicolumn{4}{c}{\textbf{True Function $f^*$}} \\
        & & abs & linear & quad &sin \\ 
        \midrule
        \multirow{3}{*}{LS} &\texttt{Silverman}    & 0.076 $\pm$ 0.032 & 0.050 $\pm$ 0.053 & 0.024 $\pm$ 0.032 & 0.124 $\pm$ 0.105 \\
        &\texttt{Med-Heuristic} & 0.100 $\pm$ 0.081 & 0.058 $\pm$ 0.053 & 0.024 $\pm$ 0.032 & 0.146 $\pm$ 0.091\\
        &\texttt{Our Method} & 0.051 $\pm$ 0.029 & 0.048 $\pm$ 0.054 & 0.025 $\pm$ 0.032 & 0.100 $\pm$ 0.111\\
        \midrule
        \multirow{3}{*}{LW} &\texttt{Silverman}    & 0.282 $\pm$ 0.067 & 0.096 $\pm$ 0.044 & 0.050 $\pm$ 0.022 & 0.169 $\pm$ 0.047 \\
        &\texttt{Med-Heuristic} & \textbf{0.127} $\pm$ \textbf{0.075} & 0.050 $\pm$ 0.032 & 0.040 $\pm$ 0.026 & 0.101 $\pm$ 0.066\\
        &\texttt{Our Method} & 0.128 $\pm$ 0.077 & \textbf{0.036} $\pm$ \textbf{0.017} & \textbf{0.026} $\pm$ \textbf{0.012} & \textbf{0.041} $\pm$ \textbf{0.022}\\
        \midrule
        \multirow{3}{*}{NS} &\texttt{Silverman}    & 0.136 $\pm$ 0.080 & 0.153 $\pm$ 0.115 & 0.071 $\pm$ 0.042 & 0.198 $\pm$ 0.147 \\
        &\texttt{Med-Heuristic} & 0.315 $\pm$ 0.260 & 0.571 $\pm$ 0.823 & 1.229 $\pm$ 2.753 & 0.560 $\pm$ 0.322\\
        &\texttt{Our Method} & \textbf{0.056} $\pm$ \textbf{0.040} & \textbf{0.029} $\pm$ \textbf{0.012} & \textbf{0.043} $\pm$ \textbf{0.042} & \textbf{0.052} $\pm$ \textbf{0.027} \\
        \bottomrule
        \end{tabular}}
        \label{tab:mse_f4_100}
    \end{table}

\subsection{More Experiment Settings}\label{sec:exp_setting}
we standardize the values of $Y$ to have zero mean and unit variance for numerical stability. Experiments on each setting are repeated 10 times. To avoid overfitting, we compute the empirical risk in the KEIC as a two-fold cross validation error. Random seed is set to 527. Codes for the experiments are provided in the supplementary material.

\end{document}